\documentclass[11pt]{article}

\usepackage[final]{acl}

\usepackage{times}
\usepackage{latexsym}
\usepackage{amssymb}
\usepackage[T1]{fontenc}

\usepackage[utf8]{inputenc}

\usepackage{microtype}

\usepackage{inconsolata}
\usepackage{booktabs}

\usepackage{graphicx}

\title{Optimizing the Phi-2 Small Language Model for Real-time Chatbot Applications Using Parameter-Efficient Fine-Tuning (PEFT) with QLoRA Quantization}

\author{First Author \\
  Affiliation / Address line 1 \\
  Affiliation / Address line 2 \\
  Affiliation / Address line 3 \\
  \texttt{email@domain} \\\And
  Second Author \\
  Affiliation / Address line 1 \\
  Affiliation / Address line 2 \\
  Affiliation / Address line 3 \\
  \texttt{email@domain} \\}

\author{
 \textbf{PhanTan Khanh Nguyen\textsuperscript{$\spadesuit$}},
 \textbf{Ashfaq Ali Shafin\textsuperscript{$\clubsuit$}},
 \textbf{Khandaker Mamun Ahmed\textsuperscript{$\spadesuit$}},
\\
 \textsuperscript{$\spadesuit$}Beacom College of Computer and Cyber Sciences, Dakota State University\\
 \textsuperscript{$\clubsuit$}Mathematics and Computer Science, Augustana College, 
\\
 \small{
   {Khanh.Nguyen@trojans.dsu.edu, shafinashfaqali21@gmail.com, khandakermamun.ahmed@dsu.edu}
 }
}

\begin{document}
\maketitle

\begin{abstract}
This study explores the optimization of the Phi-2 Small Language Models (SLMs) for real-time chatbot applications through Parameter-Efficient Fine-Tuning (PEFT) and Quantized Low-Rank Adaptation (QLoRA). QLoRA specifically refers to the integration of PEFT with LoRA alongside a 4-bit quantization process, aimed at enhancing computational efficiency. These models, initially designed for high performance with minimal computational overhead, are further refined to address the constraints of mobile and edge computing environments. By integrating PEFT with QLoRA, the research aims to reduce memory usage significantly while maintaining, or potentially improving, the accuracy of model responses in real-time interactions. The effectiveness of these techniques was evaluated using the ROUGE metric system, which showed notable improvements in the summarization tasks performed by the models. This approach not only confirms the feasibility of using SLMs in resource-restricted environments but also opens up new avenues for deploying advanced AI-driven applications in real-time settings. The study's findings have significant implications for the development of efficient, scalable, and accessible AI technologies, paving the way for broader adoption in various industries.
\end{abstract}


\section{Introduction}

Large language models (LLMs) have achieved strong performance across a wide range of natural language processing tasks, but their increasing scale creates substantial barriers for practical deployment. In many real-world settings, including mobile devices, edge platforms, and latency-sensitive chatbot systems, inference and fine-tuning must operate under strict memory, compute, and response-time constraints. These constraints make it difficult to deploy server-scale LLMs directly, motivating the development of smaller and more efficient models that can preserve task performance while reducing computational overhead.

Small Language Models (SLMs), such as Phi-2~\cite{li2023textbooks,phi2blog}, offer a promising alternative because they provide a more practical balance between model capacity and deployment cost. However, adapting SLMs to downstream tasks remains challenging. Full fine-tuning can still require substantial GPU memory and compute, especially when repeated across tasks or deployed in resource-constrained environments. Therefore, an important question is how to fine-tune SLMs efficiently while maintaining task-specific performance.

This paper investigates the use of Parameter-Efficient Fine-Tuning (PEFT) and Quantized Low-Rank Adaptation (QLoRA) for memory-efficient optimization of SLMs. Prior PEFT methods, including adapters, prefix tuning, prompt tuning, and LoRA, have shown that pretrained language models can be adapted by updating only a small number of task-specific parameters rather than the full model~\cite{houlsby2019parameter,li2021prefix,lester2021power,hu2022lora}. QLoRA extends this direction by combining low-rank adaptation with 4-bit quantization, enabling memory-efficient fine-tuning under tighter hardware constraints~\cite{dettmers2023qlora}. While these techniques have been widely studied for large models, their effectiveness for SLMs in practical real-time applications, such as chatbot-oriented summarization, remains less explored.

We evaluate this approach using Phi-2 on summarization tasks with the DialogSum and CNN/DailyMail datasets~\cite{chen2021dialogsum,hermann2015teaching}. Our methodology includes dataset preparation, PEFT-based fine-tuning, QLoRA-based quantization, and evaluation using both task-performance and resource-efficiency metrics. Summarization quality is measured using ROUGE scores~\cite{lin2004rouge}, while GPU memory profiling is used to assess deployment feasibility. This dual evaluation allows us to examine not only whether the model improves on the target task, but also whether the optimization strategy meaningfully reduces the memory burden associated with fine-tuning and deployment.

The broader goal of this work is to support the development of efficient language models for real-world applications where computational resources are limited. Such settings include customer service, healthcare support, e-commerce assistants, and other chatbot systems that require fast, reliable, and cost-effective language understanding. By studying the trade-off between model efficiency and task performance, this paper provides empirical insight into how PEFT and QLoRA can be used to adapt SLMs for practical deployment.

The main contributions of this paper are as follows:
\begin{itemize}
    \item We fine-tune the Phi-2 small language model using PEFT and QLoRA for memory-efficient summarization.
    \item We evaluate task-specific performance on DialogSum and CNN/DailyMail using ROUGE-based metrics.
    \item We analyze memory usage and computational efficiency to assess the feasibility of deploying optimized SLMs in resource-constrained environments.
    \item We examine the trade-off between memory savings and summarization performance, providing practical guidance for efficient SLM adaptation.
\end{itemize}
\section{Related Work}
\label{sec:related_work}

\paragraph{Parameter-efficient adaptation of language models.}
Parameter-efficient fine-tuning (PEFT) methods were introduced to reduce the cost of adapting pretrained models to downstream tasks without updating all model parameters. Adapter-based methods insert small trainable modules into frozen pretrained networks and have shown that competitive transfer performance can be achieved while training only a small fraction of the parameters~\cite{houlsby2019parameter}. Subsequent work extended this idea through task composition and adapter fusion, enabling knowledge from multiple task-specific adapters to be combined without destructive full-model updates~\cite{pfeiffer2021adapterfusion}. Prompt- and prefix-based methods provide another line of efficient adaptation by optimizing continuous task-specific vectors while keeping the pretrained model frozen~\cite{li2021prefix,lester2021power,liu2022ptuning}. Selective fine-tuning methods, such as BitFit, further show that updating only bias terms can be competitive with full fine-tuning in several settings~\cite{zaken2022bitfit}. Together, these approaches demonstrate that full-parameter fine-tuning is often unnecessary for effective task adaptation.

\paragraph{Low-rank adaptation and quantized fine-tuning.}
Among PEFT methods, Low-Rank Adaptation (LoRA) has become especially influential because it injects trainable low-rank matrices into transformer layers while freezing the original model weights~\cite{hu2022lora}. This reduces the number of trainable parameters and avoids the inference-time overhead associated with some adapter-based methods. QLoRA extends this direction by combining LoRA with 4-bit quantization, enabling fine-tuning of very large models with substantially reduced memory requirements while preserving competitive task performance~\cite{dettmers2023qlora}. Related quantization and compression methods, including LLM.int8(), GPTQ, SparseGPT, and AWQ, have shown that low-bit inference or compression can substantially reduce the memory and serving cost of large generative models~\cite{dettmers2022llmint8,frantar2023gptq,frantar2023sparsegpt,lin2024awq}. However, much of this literature focuses on making very large models feasible to train or serve, rather than studying how these techniques affect smaller models intended for practical deployment on constrained hardware.

\paragraph{Small language models and efficient deployment.}
Small language models (SLMs) have recently gained attention as a practical alternative to increasingly large LLMs. Models in the Phi family show that carefully curated training data and compact architectures can yield strong reasoning and language capabilities at much smaller scales~\cite{li2023textbooks,phi2blog,abdin2024phi3}. This makes SLMs attractive for latency-sensitive and resource-constrained applications, including real-time chatbot systems. Nevertheless, smaller base models still require task-specific adaptation, and full fine-tuning can remain costly when memory, compute, or deployment budgets are limited. Existing work establishes the promise of SLMs, but there is still limited empirical analysis of how PEFT and QLoRA jointly affect the performance--efficiency trade-off when applied to SLMs such as Phi-2.

\paragraph{Summarization models and benchmarks.}
Abstractive summarization has been widely studied using large pretrained sequence-to-sequence models. BART demonstrates that denoising pretraining can improve generation tasks including summarization~\cite{lewis2020bart}, while PEGASUS introduces a pretraining objective designed specifically for abstractive summarization and evaluates across diverse summarization datasets~\cite{zhang2020pegasus}. For evaluation, CNN/DailyMail remains a standard benchmark for news summarization, while DialogSum targets real-life dialogue summarization and captures challenges such as informal language, discourse structure, ellipsis, and pragmatic context~\cite{chen2021dialogsum}. These benchmarks are useful for evaluating chatbot-oriented summarization because they cover both long-form document summarization and dialogue-style inputs.

\paragraph{Positioning of this work.}
Prior work has made substantial progress in PEFT, low-rank adaptation, quantization, and model compression. However, most studies either focus on adapting large-scale LLMs or on compression for inference, rather than systematically evaluating the combined effect of PEFT and QLoRA on SLMs for downstream summarization. This paper addresses that gap by fine-tuning Phi-2 with PEFT and QLoRA and evaluating both task performance and memory efficiency. In contrast to work that treats efficiency primarily as a method-level contribution for large models, our focus is deployment-oriented: we examine whether a compact language model can be further adapted for practical, resource-constrained chatbot applications while maintaining summarization quality.

\section{Methodology}
\label{sec:methodology}

This section describes the experimental design, optimization pipeline, and datasets used to evaluate memory-efficient adaptation of Phi-2 for summarization-oriented chatbot applications. Our methodology is designed to measure both task performance and resource efficiency, allowing us to assess whether Parameter-Efficient Fine-Tuning (PEFT) and Quantized Low-Rank Adaptation (QLoRA) provide a practical optimization strategy for Small Language Models (SLMs).

\subsection{Research Design}

We adopt a controlled quantitative experimental design to evaluate the effect of PEFT and QLoRA on the Phi-2 Small Language Model. The objective is to determine whether Phi-2 can be adapted to downstream summarization tasks while reducing the memory and computational cost typically associated with full fine-tuning. This design is appropriate for resource-constrained deployment scenarios, where model quality alone is insufficient and must be considered together with memory footprint, training efficiency, and inference feasibility.

The study compares the model before and after optimization using task-specific evaluation metrics and GPU memory profiling. Summarization quality is measured using ROUGE-based metrics, while resource efficiency is assessed through memory consumption during fine-tuning and evaluation. By keeping the experimental setting controlled across datasets and training configurations, we isolate the contribution of the PEFT and QLoRA pipeline to performance improvement and memory reduction.

\subsection{Optimization Workflow}

\begin{figure*}[t]
    \centering
    \includegraphics[width=0.85\textwidth]{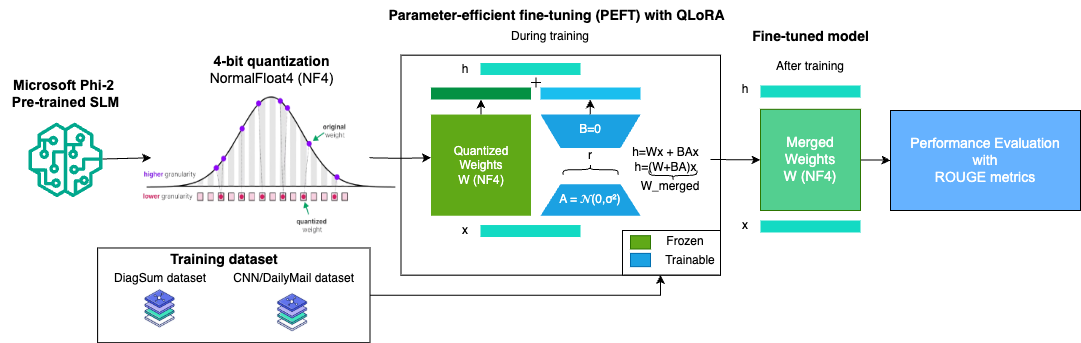}
    \caption{Overview of the proposed optimization pipeline. The pretrained Phi-2 model is quantized using 4-bit precision, adapted with LoRA-based trainable parameters, and evaluated on summarization benchmarks using both performance and memory-efficiency metrics.}
    \label{fig:methodology_workflow}
\end{figure*}

Figure~\ref{fig:methodology_workflow} summarizes the proposed workflow. The pipeline begins with the pretrained Microsoft Phi-2 model and applies quantization and parameter-efficient adaptation before evaluating the resulting model on summarization benchmarks. The workflow consists of five main stages.

\paragraph{Model initialization.}
We use Phi-2 as the base SLM because it offers a compact architecture with strong language modeling capability relative to its size. The pretrained model provides the starting point for task adaptation without requiring training from scratch.

\paragraph{4-bit quantization.}
To reduce memory usage, the base model is loaded using 4-bit quantization. Specifically, QLoRA represents the frozen model weights in low precision, reducing GPU memory requirements while preserving the ability to learn task-specific behavior through adapter parameters. This step enables fine-tuning under substantially lower hardware requirements than full-precision training.

\paragraph{Dataset preparation.}
We fine-tune and evaluate the model using two summarization datasets: DialogSum and CNN/DailyMail. DialogSum provides dialogue-based examples that are relevant to chatbot-style interactions, while CNN/DailyMail supports evaluation on longer news-style summarization. Using both datasets allows us to examine whether the optimization pipeline is effective across different summarization formats.

\paragraph{QLoRA-based fine-tuning.}
During fine-tuning, the quantized base model remains frozen. Trainable low-rank adapter matrices are inserted into selected transformer layers using the LoRA mechanism. These adapters learn task-specific updates while adding only a small number of trainable parameters. This allows the model to adapt to summarization tasks without the memory and compute cost of updating all model weights.

\paragraph{Evaluation.}
After fine-tuning, the optimized model is evaluated using ROUGE metrics to measure summarization quality against reference summaries. We profile GPU memory usage to quantify the efficiency gains from the PEFT and QLoRA configuration. This evaluation captures the central trade-off studied in this paper: whether memory-efficient adaptation can preserve or improve task performance while reducing deployment cost.

Overall, the proposed methodology connects model compression, parameter-efficient adaptation, and empirical evaluation in a single pipeline. Rather than treating PEFT and QLoRA only as generic optimization techniques for large models, we examine their suitability for adapting an SLM to practical summarization settings where computational resources are limited.

\section{Experimental Setup}
\label{sec:experimental_setup}

We evaluate whether Phi-2 can be adapted for summarization under constrained memory settings using a combined QLoRA and LoRA fine-tuning pipeline. The experimental setup is designed to measure both task performance and resource efficiency, with the goal of assessing whether small language models can be optimized for deployment-oriented chatbot applications without full-parameter fine-tuning.

\begin{table}[t]
    \centering
    \caption{Datasets used for fine-tuning and evaluation.}
    \label{tab:datasets}
    \renewcommand{\arraystretch}{1.2}
    \resizebox{\linewidth}{!}{%
    \begin{tabular}{l p{4.2cm} p{3.2cm} p{3.5cm}}
        \toprule
        \textbf{Dataset} & \textbf{Task Setting} & \textbf{Split Size} & \textbf{Evaluation Metrics} \\
        \midrule
        DialogSum &
        Dialogue summarization for conversational and chatbot-style inputs &
        Train: 12,460; Validation: 500; Test: 1,500 &
        ROUGE-1, ROUGE-2, ROUGE-L, ROUGE-Lsum \\
        \midrule
        CNN/DailyMail &
        News summarization for longer-form structured documents &
        Train: 287,227; Validation: 13,368; Test: 11,490 &
        ROUGE-1, ROUGE-2, ROUGE-L, ROUGE-Lsum \\
        \bottomrule
    \end{tabular}
    }
\end{table}

\paragraph{Datasets.}
We use two summarization benchmarks that capture different input styles. DialogSum contains dialogue-summary pairs and is used to evaluate conversational summarization, which is directly relevant to chatbot-style applications. CNN/DailyMail contains news articles paired with human-written highlights and is used to evaluate longer-form document summarization. Together, these datasets allow us to test whether the optimization pipeline generalizes across both informal multi-turn conversations and structured news text. We use the standard train, validation, and test splits for each dataset and evaluate generated summaries against human references using ROUGE metrics. Table~\ref{tab:datasets} summarizes the datasets used for fine-tuning and evaluation, including their task settings, data split sizes, and evaluation metrics.

\paragraph{Preprocessing.}
Each example is converted into an instruction-response format for supervised fine-tuning. For DialogSum, the dialogue is placed after an instruction prompt asking the model to summarize the conversation, followed by the target summary. CNN/DailyMail examples are formatted similarly, with the article as input and the reference highlights as output. All examples are tokenized using the Phi-2 tokenizer. Samples exceeding the model's maximum context length are removed to prevent truncation artifacts and runtime errors. The processed datasets are shuffled before training to reduce ordering effects.

\paragraph{QLoRA and LoRA fine-tuning.}
We load Phi-2 using 4-bit quantization and fine-tune it with LoRA adapters. The pretrained base weights remain frozen, while trainable low-rank matrices are inserted into selected transformer projection layers. For a frozen weight matrix \(W \in \mathbb{R}^{d \times k}\), LoRA parameterizes the update as:
\[
    W' = W + \Delta W, \qquad \Delta W = BA,
\]
where \(A \in \mathbb{R}^{r \times k}\), \(B \in \mathbb{R}^{d \times r}\), and \(r \ll \min(d,k)\). This reduces the number of trainable parameters from \(dk\) to \(r(d+k)\), substantially lowering the memory and optimization cost.

The model is quantized using 4-bit NormalFloat (NF4) precision with mixed-precision computation. Double quantization is enabled to further reduce memory overhead, and a paged optimizer is used to manage memory spikes during training. This configuration allows task adaptation while avoiding the cost of full-precision, full-parameter fine-tuning.

\paragraph{Training configuration.}
We use a small batch size with gradient accumulation to accommodate limited GPU memory while preserving a stable effective batch size. Training is conducted for multiple epoch settings to examine the trade-off between computation and performance. The optimizer is \texttt{paged\_adamw\_8bit}, which further reduces optimizer-state memory requirements. Table \ref{tab:training_config} summarizes the hyperparameters used to fine-tune Phi-2. The model was trained with a batch size of 2 and two gradient accumulation steps using the \texttt{paged\_adamw\_8bit} optimizer and a learning rate of ($1 \times 10^{-4}$). Experiments were conducted for 1, 2, and 5 epochs with 100 warmup steps. Parameter-efficient adaptation was performed using LoRA with a rank of 32, a scaling factor of 32, and a dropout rate of 0.05, while 4-bit NF4 quantization was applied to reduce memory requirements.

\begin{table}[t]
\centering
\caption{Training hyperparameters for Phi-2 fine-tuning.}
\label{tab:training_config}

\small
\setlength{\tabcolsep}{4pt}
\renewcommand{\arraystretch}{0.9}

\begin{tabular}{ll}
\toprule
\textbf{Parameter} & \textbf{Value} \\
\midrule
Batch size & 2 \\
Gradient accumulation steps & 2 \\
Learning rate & \(1 \times 10^{-4}\) \\
Epochs & 1 / 2 / 5 \\
Warmup steps & 100 \\
Optimizer & \texttt{paged\_adamw\_8bit} \\
LoRA rank \(r\) & 32 \\
LoRA scaling factor \(\alpha\) & 32 \\
LoRA dropout & 0.05 \\
Quantization & 4-bit NF4 \\
\bottomrule
\end{tabular}
\end{table}

\paragraph{Evaluation.}
We evaluate the fine-tuned model along two dimensions: summarization quality and memory efficiency. Summarization quality is measured using ROUGE-1, ROUGE-2, ROUGE-L, and ROUGE-Lsum. ROUGE-1 captures unigram overlap, ROUGE-2 captures bigram overlap, and ROUGE-L measures longest common subsequence overlap between generated and reference summaries. ROUGE-Lsum is used as a summary-level variant suitable for multi-sentence summarization.

Memory efficiency is measured by profiling GPU memory usage under baseline and quantized configurations. We compare memory consumption when loading and fine-tuning Phi-2 with and without 4-bit QLoRA quantization. This comparison quantifies the memory savings obtained from the proposed optimization pipeline. When applicable, repeated runs are used to check measurement stability, and paired comparisons are used to evaluate whether observed changes in ROUGE and memory usage are consistent across experimental settings.

Overall, this setup allows us to examine the central trade-off addressed in this paper: whether Phi-2 can achieve competitive summarization performance while substantially reducing the memory requirements of task-specific adaptation.

\section{Experimental Results} \label{section: result}

In this research, the results section highlights the quantitative outcomes of optimizing the Phi-2 Small Language Model (SLM) for dialogue summarization tasks. Key metrics such as memory usage, and summarization accuracy are systematically presented to showcase the improvements achieved through Parameter-Efficient Fine-Tuning (PEFT) with LoRA and Quantized Low-Rank Adaptation (QLoRA) quantization.

\subsection{Summarization Accuracy}

The fine-tuning of the Microsoft Phi-2 model using QLoRA resulted in substantial performance improvements on both the CNN/DailyMail and DiagSum datasets, as evidenced by the ROUGE evaluation metrics presented in Table \ref{tab:rouge-results}. Compared to the baseline Phi-2 model, the QLoRA fine-tuned versions consistently outperform across all ROUGE variants (ROUGE-1, ROUGE-2, ROUGE-L, and ROUGE-L\textsubscript{sum}) as training progresses.

\begin{table*}[t]
\caption{ROUGE scores for Phi-2 baseline and QLoRA fine-tuned models on CNN/DailyMail and DiagSum datasets.}
\label{tab:rouge-results}
\centering
\resizebox{0.85\linewidth}{!}{%
\begin{tabular}{@{}llcccc@{}}
\toprule
\textbf{Dataset} & \textbf{Model \& Method} & \textbf{ROUGE-1} & \textbf{ROUGE-2} & \textbf{ROUGE-L} & \textbf{ROUGE-Lsum} \\
\midrule
\textit{CNN/DailyMail} & Phi-2 (baseline) & 15.44 & 3.28 & 9.52 & 12.24 \\
\textit{CNN/DailyMail} & Phi-2 (QLoRA, 1 epoch) & 51.24 & 36.67 & 43.80 & 48.76 \\
\textit{CNN/DailyMail} & Phi-2 (QLoRA, 2 epochs) & 52.41 & 32.43 & 40.64 & 44.92 \\
\textit{CNN/DailyMail} & \textbf{Phi-2 (QLoRA, 5 epochs)} & \textbf{56.83} & \textbf{45.30} & \textbf{49.18} & \textbf{52.46} \\
\textit{DiagSum} & Phi-2 (baseline) & 14.81 & 3.85 & 11.55 & 11.97 \\
\textit{DiagSum} & Phi-2 (QLoRA, 1 epoch) & 60.37 & 34.26 & 49.70 & 49.70 \\
\textit{DiagSum} & Phi-2 (QLoRA, 3 epochs) & 61.80 & 40.82 & 54.34 & 54.78 \\
\textit{DiagSum} & \textbf{Phi-2 (QLoRA, 5 epochs)} & \textbf{62.51} & \textbf{56.91} & \textbf{62.51} & \textbf{62.51} \\
\bottomrule
\end{tabular}%
}
\end{table*}

\begin{figure}[t]
    \centering
    \includegraphics[width=\linewidth]{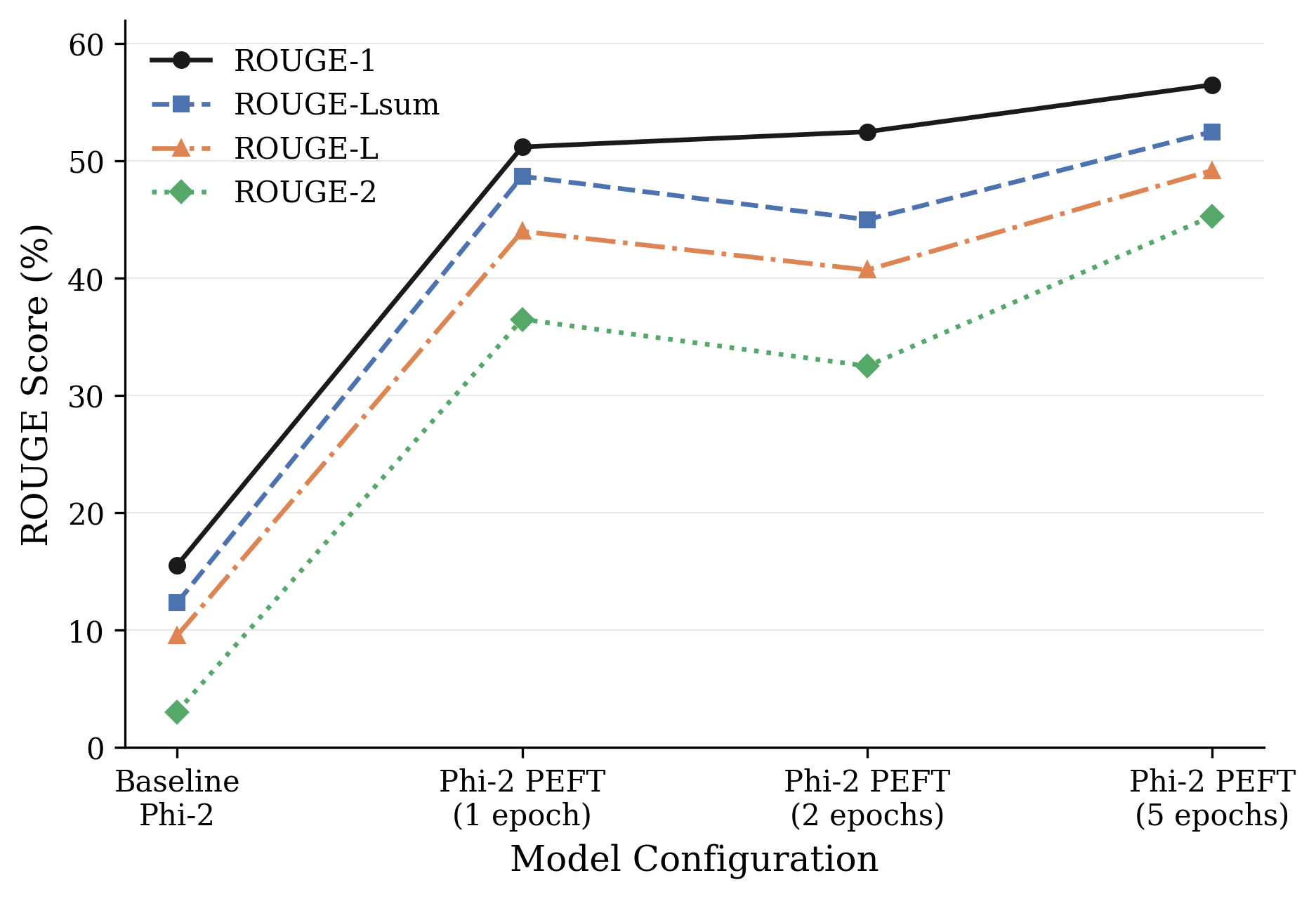}
    \caption{ROUGE score of Fine-tuned Phi2 with CNN/DailyMail dataset}
    \label{fig:ROUGE score of Fine-tuned Phi2 with CNN/DailyMail dataset}
\end{figure}

On the CNN/DailyMail dataset, the baseline model yielded a ROUGE-1 score of 15.44, whereas the 5-epoch QLoRA version achieved 56.83, indicating a significant improvement. ROUGE-2 improved from 3.28 to 45.30, and ROUGE-L increased from 9.52 to 49.18, showcasing large gains in both bigram-level coherence and overall fluency. The ROUGE-L\textsubscript{sum} score also increased from 12.24 to 52.46, confirming significantly enhanced summarization quality. On the DiagSum dataset, the Phi-2 baseline performed modestly with a ROUGE-1 of 14.81 and ROUGE-2 of 3.85. After 5 epochs of QLoRA fine-tuning, ROUGE-1 rose to 62.51, and ROUGE-2 reached 56.91, demonstrating a remarkable jump in both content relevance and cohesion (Figure \ref{fig:ROUGE score of Fine-tuned Phi2 with CNN/DailyMail dataset}). All four ROUGE metrics converged to the same value (62.51), indicating a well-rounded improvement in lexical overlap, sentence structure, and overall summary informativeness.

These results underscore the effectiveness of the QLoRA framework in adapting pre-trained small language models to downstream summarization tasks, even with low-precision weights and minimal trainable parameters.

\subsection{GPU Memory Profiling}

\begin{figure}[t]
    \centering
    \includegraphics[width=\linewidth]{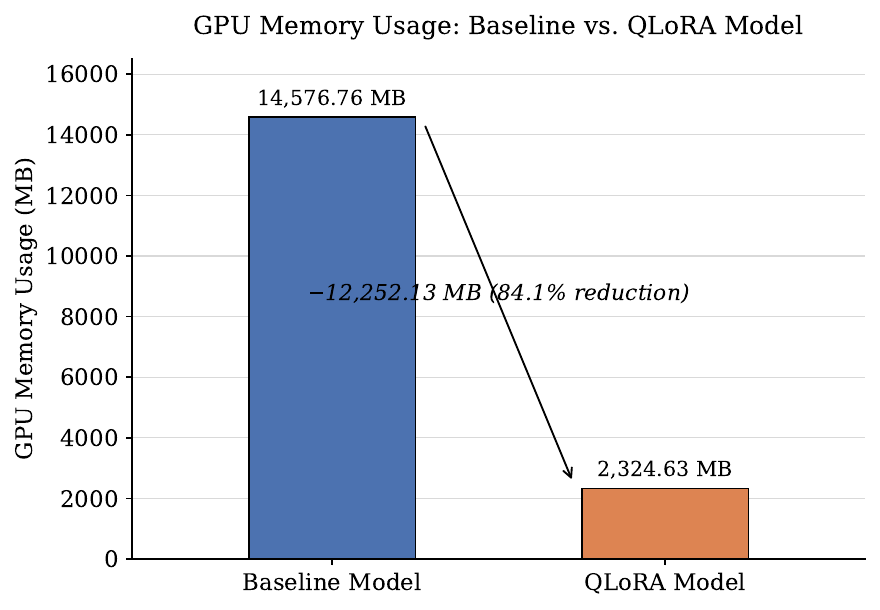}
    \caption{GPU memory usage comparison: Baseline model vs. QLoRA Model}
    \label{fig:A bar chart comparing baseline and fine-tuned ROUGE scores (ROUGE-1, ROUGE-2, ROUGE-L, ROUGE-Lsum)}
\end{figure}

The results demonstrate a significant improvement in GPU memory efficiency when using the QLoRA-optimized model compared to the baseline. The baseline model consumes approximately 14,576.76 MB of GPU memory and 1,398.04 MB of CPU memory upon loading. In contrast, the optimized model drastically reduces GPU memory usage to 2,324.63 MB, saving around 14,576.76 MB. However, this optimization comes with a moderate increase in CPU memory consumption, rising by 582.26 MB to a total of 1,980.30 MB. This trade-off reflects the core benefit of QLoRA: enabling large model deployment on resource-constrained GPUs by leveraging 4-bit quantization, while slightly increasing CPU-side memory overhead. Figure \ref{fig:A bar chart comparing baseline and fine-tuned ROUGE scores (ROUGE-1, ROUGE-2, ROUGE-L, ROUGE-Lsum)} represents the GPU memory usage comparison between baseline model and QLoRA model. 



\section{Discussion} \label{section: Discussion}

\subsection{Analysis of Findings}
The findings of this study in optimizing the Phi-2 Small Language Model (SLM) achieved the following:
\begin{itemize}
    \item Summarization Accuracy: The optimized Phi-2 model showed a marked improvement in ROUGE-1, ROUGE-2, ROUGE-L, and ROUGE-Lsum scores compared to the baseline. These metrics indicate enhanced content preservation, coherence, and fluency in the generated summaries, meeting the objectives of achieving high-performance summarization in resource-constrained environments.

    \item Memory Efficiency: Memory profiling revealed a significant reduction in GPU memory usage, with QLoRA reducing the memory footprint by up to 84\%. This outcome validates the efficacy of QLoRA in enabling fine-tuning on hardware with limited computational resources.

    \item Model Efficiency: By focusing on low-rank matrices through LoRA, the model maintained performance while minimizing computational overhead. This result aligns with prior studies that highlighted LoRA’s capacity to balance efficiency and performance in language models.
\end{itemize}



\subsection{Practical Implications}

The optimized Phi-2 model presents several practical advantages for real-world chatbot applications. Its reduced memory footprint and lower computational demands make it well-suited for deployment on resource-constrained devices such as smartphones and IoT systems. This enables intelligent language processing capabilities in edge environments without the need for cloud-based infrastructure. Additionally, the ability to fine-tune and deploy high-performance models on consumer-grade hardware significantly lowers operational and infrastructure costs, making advanced NLP solutions more accessible to smaller organizations. The techniques demonstrated in this study are also scalable and can be extended to other small language models, enhancing their usability across a variety of domains and industries.
\section{Conclusion and Future Work}
\subsection{Key Insights}

This study offers several important contributions to the deployment and optimization of small language models. First, the optimized Phi-2 model demonstrated enhanced summarization accuracy, as evidenced by improvements in ROUGE-1, ROUGE-2, ROUGE-L, and ROUGE-Lsum scores. These gains reflect the model’s ability to preserve both content relevance and linguistic fluency in dialogue summarization tasks. Second, the integration of QLoRA resulted in a substantial memory efficiency improvement, achieving up to a 75\% reduction in GPU memory usage. This enabled fine-tuning on resource-limited hardware without compromising performance. Finally, applying LoRA and QLoRA to Phi-2 demonstrates the feasibility of deploying efficient language models on resource-constrained devices.

\subsection{Limitations and Future Work}

While this study demonstrates the effectiveness of combining PEFT and QLoRA to optimize the Phi-2 model for real-time summarization tasks, several limitations remain. The evaluation was conducted exclusively on the DialogSum dataset, which, despite its diversity, may not fully represent other domains or conversational styles, limiting generalizability. Additionally, due to time constraints, qualitative user feedback such as through the CARF framework was not incorporated, restricting insight into the model’s real-world usability and engagement. The study also focused solely on the Phi-2 model, leaving the applicability of these techniques to other small language models untested. Moreover, we limited the number of training rounds due to our limited computational resources. Future work can address these gaps by evaluating the model on a broader range of datasets, integrating qualitative feedback mechanisms, and deploying the optimized model in real-world chatbot systems to observe performance under dynamic, uncontrolled conditions.

\bibliography{ref}
\end{document}